\documentclass[11pt,a4paper]{article}
\usepackage[hyperref]{emnlp2020}
\usepackage{times}
\usepackage{latexsym}

\usepackage{microtype}

\aclfinalcopy 

\usepackage{amsmath,amssymb,amsfonts}
\usepackage{url,enumitem}
\usepackage{booktabs}
\usepackage{xspace}
\usepackage{graphicx}
\usepackage{subcaption}
\usepackage{comment}
\usepackage[normalem]{ulem}
\newcommand{\Sref}[1]{\S\ref{#1}}
\DeclareMathOperator{\argmin}{argmin}

\title{Fortifying Toxic Speech Detectors Against Veiled Toxicity}
\author{Xiaochuang Han \\
  Carnegie Mellon University \\
  \texttt{xiaochuh@cs.cmu.edu} \\\And
  Yulia Tsvetkov \\
  Carnegie Mellon University \\
  \texttt{ytsvetko@cs.cmu.edu} \\}

\date{}

\begin{document}
\maketitle
\begin{abstract}
Modern toxic speech detectors are incompetent in recognizing disguised offensive language, such as adversarial attacks that deliberately avoid known toxic lexicons, or manifestations of implicit bias. Building a large annotated dataset for such veiled toxicity can be very expensive. In this work, we propose a framework aimed at fortifying existing toxic speech detectors without a large labeled corpus of veiled toxicity. Just a handful of probing examples are used to surface orders of magnitude more disguised offenses. We augment the toxic speech detector's training data with these discovered offensive examples, thereby making it more robust to veiled toxicity while preserving its utility in detecting overt toxicity.\footnote{Our code is available at \url{https://github.com/xhan77/veiled-toxicity-detection}.}
\\
\underline{Warning}: this paper contains examples that may be offensive or upsetting.
\end{abstract}

\section{Toxic Language in Disguise}
\label{sec:intro}

Toxic language has been recognized as a severe problem in the online social communities.
While great efforts have been made to detect and prevent the spread of overt trolling, hate speech, abusive language, and toxic comments \citep{Schmidt2017ASO,fortuna2018survey},
they often build upon lexicon-based approaches \citep{Waseem2016HatefulSO,Davidson2017AutomatedHS} and thus are ineffective at detecting forms of veiled toxicity; e.g., codewords \citep{Taylor2017SurfacingCH}, novel forms of offense \citep{Jain2018AdversarialTG}, and subtle and often unintentional manifestations of social bias such as microaggressions and condescension \citep{Breitfeller2019FindingMI,Wang2019TalkDownAC}.

In this work, we focus on disguised toxic language that is often undetected by existing tools.\footnote{We use the terms \emph{veiled} and \emph{disguised} toxicity interchangeably in this work.} It can be innocuous on the surface lexicon-level, but as offensive as hate speech and lastingly harmful \citep{Sue2007RacialMI,Sue2010MicroaggressionsIE,Nadal2014TheIO,Jurgens2019AJA}.\footnote{To demonstrate, an offensive example from the Social Bias Inference Corpus \citep{Sap2019SocialBF} \emph{``What is the difference between a bench and an African? A bench can support a family.''} is determined as non-toxic by a popular toxic speech detector Perspective API, \url{https://perspectiveapi.com/}; the example gets a toxicity score of 0.1 out of 1. It is originally from Reddit's r/darkJokes.}

Detecting veiled toxicity is hard: deep semantic analysis and large datasets are needed. But veiled offenses are not represented in existing toxicity datasets \citep{Waseem2016HatefulSO,Davidson2017AutomatedHS,Founta2018LargeSC} and building a new dataset is expensive: candidates for annotation cannot be filtered through lexicons and random sampling of social media posts will surface only a tiny fraction of relevant examples \citep{Breitfeller2019FindingMI}. Moreover, since biased text is often unconscious and subjective, untrained annotators might mislabel it due to their own biases \cite{Breitfeller2019FindingMI, field2020unsupervised}.

We propose a framework to surface veiled offenses and improve toxicity classifiers that are compromised in detecting them. It requires a small set of labeled probing examples
to surface orders of magnitude more disguised offenses missed by the classifier, through interpretable ML techniques tracking the \emph{influence} of training examples on classifier's predictions \citep{Koh2017UnderstandingBP,Pruthi2020EstimatingTD}.
From an original classifier that detects veiled toxicity with an accuracy of 1\%, we achieve up-to 51\% accuracy in detecting veiled offenses while preserving the utility of the classifier in detecting overt offenses.
To the best of our knowledge, our work is the first in making toxic speech detectors robust against veiled toxicity with almost no annotated data.

\section{Identifying Veiled Toxicity}
A typical toxicity classifier $C$ might fail to identify veiled offenses because they are not well represented among toxic examples in its training data $\mathcal{D}$. Moreover, the non-toxic portion of $\mathcal{D}$ might be polluted with (mislabeled) disguised offenses.
At inference time, $C$ might thus mislabel, for example, microaggressions as well as adversarial attacks deliberately avoiding known toxic lexicons.

To make such compromised classifier $C$ more robust, we propose a framework that does not require access to $\mathcal{D}$.\footnote{Since hate speech training data is often proprietary and platform-specific \cite{macavaney2019hate}.}
We start with a dataset $\mathcal{D'}$ comprising examples that can be labeled as offensive or non-offensive.\footnote{Here we assume a binary classifier for simplicity, but our setup is applicable to multi-class settings.}
We build a student model $C'$ on $\mathcal{D'}$ that approximates the behavior of $C$:
$C' = \argmin_\theta \mathcal{L}(\theta, x, C(x))$, where $x$ are instances from $\mathcal{D'}$.

An effective $C'$ would mimic the behavior of $C$, including mislabeling disguised toxic examples $x_{veiled}$ as non-offensive. To address this issue, we surface the unknown $x_{veiled}$ out of the training set by probing the $C'$ with a small labeled held-out set of veiled offenses $\mathcal{P}$,\footnote{Although it could be very expensive to create large-scale datasets for disguised offenses, it is feasible to collect a very small number (e.g., less than 100) of exemplars, using methods like user reporting, e.g., \url{https://www.microaggressions.com/}.}
and tracing the model's decisions back to training examples in $\mathcal{D'}$.
We hypothesize the `influential' training examples that lead to $C'$'s predictions on $\mathcal{P}$ are likely to be $x_{veiled}$.
One probing example can surface multiple influential examples in the training data. These influential examples can then be re-annotated, making the model more robust to future veiled toxicity.

The key observation is that sampling on random a subset of non-offensive data to annotate it for disguised offensiveness is prohibitively expensive, as only a tiny fraction of annotated examples will be indeed offensive. Our proposed approach increases the likelihood to identify disguised offensiveness by surfacing training examples that influence the decisions of $C'$ when tested on $\mathcal{P}$.
The key challenge is to devise a method to track examples influential to classifier's predictions; we discuss candidate approaches in the rest of this section.

\subsection{Probing for veiled toxicity}
\label{sec:background}

We explore several methods to define the influence $\mathcal{I}(x_{trn}, x_{prb})$ of a training example $x_{trn} \in \mathcal{D'}$ over a probing example $x_{prb} \in \mathcal{P}$.

\paragraph{Embedding product}
Modern neural classifiers often consist of two parts: an encoding module $f_{enc}(\cdot)$ that transforms the input to some hidden representation, and a projection layer $f_{proj}(\cdot)$ that projects the output of the encoding module to the label space. Our first influence measure is based on the intuition that the training example with the closest embedding to the probing example in the hidden encoding space could be the most influential:
\begin{align*}
    \mathcal{I}(x_{trn}, x_{prb}) = f_{enc}(x_{trn}) \cdot f_{enc}(x_{prb}) \text{.}
\end{align*}

\paragraph{Influence functions}
\citet{Koh2017UnderstandingBP} propose influence functions for ML models, following the vision from robust statistics. It first approximates how upweighting a particular training example $(x_{trn}, y_{trn})$ in the training set $\{(x_1, y_1), \ldots, (x_n, y_n)\}$ by an infinitesimal $\epsilon_{trn}$ would change the learned model parameters $\theta$:
\begin{align*}
    \frac{d\theta}{d\epsilon_{trn}} = - H_{\theta}^{-1} \nabla_{\theta}{\mathcal{L}(\theta, x_{trn}, y_{trn})} \text{,}
\end{align*}
where $H_{\theta} = \frac{1}{n} \sum_{i=1}^{n} \nabla_{\theta}^{2}{\mathcal{L}(\theta, x_i, y_i)}$ is the Hessian of the model. We can then use the chain rule to measure how this change in the model parameters would in turn affect the loss of the probing input:
\begin{align*}
    \frac{d\mathcal{L}(\theta, x_{prb}, \hat{y}_{prb})}{d\epsilon_{trn}} = \nabla_{\theta}{\mathcal{L}(\theta, x_{prb}, \hat{y}_{prb})} \cdot \frac{d\theta}{d\epsilon_{trn}} \text{,}
\end{align*}
where $\hat{y}_{prb}$ is the \emph{wrong} label for the probing example, since we want to know which training examples lead to a \emph{wrong} prediction of the probing disguised offense. The final influence of a train example to a probing example is defined as:
    $\mathcal{I}(x_{trn}, x_{prb}) = - \frac{d\mathcal{L}(\theta, x_{prb}, \hat{y}_{prb})}{d\epsilon_{trn}}$.
More details of influence functions and their applications in NLP can be found in \citet{Koh2017UnderstandingBP} and \citet{Han2020ExplainingBB}.\footnote{Implementation details can be found in the Appendix.}

\paragraph{Gradient product}
Computing the inverse Hessian $H_{\theta}^{-1}$ in the influence functions is expensive and requires approximations if the model is non-convex. If we ignore the inverse Hessian term, the calculation reduces to the dot product between the gradient of the training loss $\mathcal{L}(\theta, x_{trn}, y_{trn})$ and the gradient of the probing loss $\mathcal{L}(\theta, x_{prb}, \hat{y}_{prb})$. This method is discussed in \citet{Pruthi2020EstimatingTD}. Specifically, we adopt the \emph{TrackIn} method, which defines the influence as:
\begin{align*}
    \mathcal{I}(x_{trn}, x_{prb}) = \sum_{i=1}^{k} \nabla_\theta \mathcal{L}(\theta_i, x_{trn}, y_{trn})\\
    \cdot \nabla_\theta \mathcal{L}(\theta_i, x_{prb}, \hat{y}_{prb})
    \text{,}
\end{align*}
where $\theta_i$ is the checkpoint of the model at each training epoch. The intuition behind this method is to approximate the total reduction in the probing loss $\mathcal{L}(\theta, x_{prb}, \hat{y}_{prb})$ during the training process when the training example $x_{trn}$ is used. More details on \emph{TrackIn} can be found in \citet{Pruthi2020EstimatingTD}.\footnote{More influence metrics can be found in \citet{Yeh2018RepresenterPS}, \citet{Khanna2019InterpretingBB}, and \citet{Barshan2020RelatIFIE}. We leave the exploration of them in our framework to future work.}

\paragraph{Training loss}
Our last method to define the influence of training examples can be considered as a baseline which is often used in  active learning as `uncertainty-based sampling' \citep{Lewis1994ASA,Zhu2008LearningAS}. The intuition is that a training example with a high loss (low confidence) could indicate that the model struggles to predict that example correctly. This alone can show that the outlier training example has a dubious label, regardless of its relationship to the probing example. For consistency, we define the influence of a training example to the mis-prediction of disguised offenses as:
    $\mathcal{I}(x_{trn}) = \mathcal{L}(\theta, x_{trn}, y_{trn})$.\footnote{Note that this method does not require probing examples.}

\section{Experiments}
\label{sec:experiments}

\subsection{Setup}
\label{sec:model_and_data_instantiation}
We use a popular toxic language detection tool, Perspective API by Jigsaw and Google, as the compromised classifier $C$. It builds upon a convolutional neural network with pretrained word embeddings and proprietary large labeled data. For the student model $C'$, we use a BERT-based model \citep{Devlin2019BERTPO}, initialized with the pretrained weights and fine-tuned on our training set. Below we instantiate our training set $\mathcal{D'}$ and a probing set $\mathcal{P}$ of veiled offenses.

\textbf{SBIC} -- Social Bias Inference Corpus  \citep{Sap2019SocialBF} is a dataset containing 45K social media posts with crowdsourced annotations of offensiveness, intention, and targeted group from a variety of origins, including hate speech, offensive language, and microaggressions,
and selected dangerous threads on Reddit (e.g., r/darkJokes) and hate sites. We use SBIC as our base dataset.
We consider three attributes in SBIC posts: offensive, target some marginalized groups, while subtly expressed. Each post's offensiveness scores can be 0 (harmless), 0.5 (maybe offensive), or 1 (offensive). We select the posts with an average offensiveness $> 0.5$  (i.e., more than half of the annotators thought it was offensive). SBIC also asks annotators to identify the potential groups of people that might be offended by the post. We keep the posts with at least one identified target group.

We first extract \textbf{veiled toxicity set}. We randomly sample 10K general reddits from no specific domains and measure their average Perspective API toxicity score $tox_{general} \approx 0.17$ on a scale [0,1]. We then measure the Perspective API toxicity scores of the posts in SBIC that are offensive to at least one minority group. We sort these scores from low to high in $x_{{1}}, x_{(2)}, \cdots, x_{(n)}$. We pick the least toxic $m$ posts as our veiled offensive language set so that $\frac{1}{m} \sum_{i=1}^{m} tox(x_{(i)}) = tox_{general}$. The extracted veiled offenses are equally non-toxic as some random general-domain reddits according to Perspective API.\footnote{All toxicity scores are under 0.5, i.e.,  Perspective API labels none of them as offensive. 12.6\% of instances with more than one annotation do not have a full annotator agreement that the instance is offensive.} There are about 3K resulting posts. We use 2K in our training set $\mathcal{D'}$ (as discussed in \Sref{sec:background} they are (mis)labeled as non-toxic), 1K for the test set, and reserve 100 for the probing set $\mathcal{P}$.

It is worth noting that Perspective API can misclassify toxic inputs for several reasons. First, although it was trained on comments from online forums such as discussions of Wikipedia and New York Times, it could misclassify SBIC examples due to a domain mismatch leading to different manifestations of overt toxicity. In addition, as we discuss above, misclassifications can be attributed to novel lexicons of toxicity, to (intentional or unintentional) spelling variations, or to more subtly expressed implicit offenses.  Our set of veiled offenses covers any of these forms, as they are hidden from the original toxicity detection model.\footnote{In practice, we observe the extracted veiled offenses to be indeed \emph{covertly} toxic, as shown in a list of randomly selected examples in the Appendix, and not a mere domain mismatch.}

We extract all SBIC posts annotated as non-offensive and also sort their Perspective API toxicity scores from low to high in $x_{{1}}, x_{(2)}, \cdots, x_{(n')}$. We pick the least toxic $m'$ posts as our \textbf{non-offensive clean set} so that $\frac{1}{m'} \sum_{i=1}^{m'} tox(x_{(i)}) = tox_{general}$. The intuition is to create a control set for the veiled offenses: these clean data are from the same domain as the veiled offenses, have the same Perspective API score as the veiled offenses on average, while being annotated in SBIC as non-offensive. There are about 10K posts in this category. We use 8K in $\mathcal{D'}$ and 1K for testing.

For the SBIC posts that are identified as offensive, we extract those with Perspective API toxicity score $>0.8$ (a recommended threshold by Perspective API for determining bad language) and consider them as our \textbf{overtly offensive set}. From 3K such posts, we use 2K in $\mathcal{D'}$ and 1K for testing.

\subsection{Student model evaluation}
Would a vanilla toxicity classifier recognize veiled offenses? We apply our student model $C'$ on the three sets in \Sref{sec:model_and_data_instantiation}. The model attains a class recall of 99.6\% and 97.2\% on the non-offensive and overtly offensive test sets, respectively. However, it fails to recognize test veiled offenses as offensive, yielding an 1.2\% class recall. In sum, it mimics Perspective API's predictions accurately.

\subsection{Evaluating probing and re-annotation}
We hypothesized that the more influential a training example is to a wrongly predicted example from $\mathcal{P}$, the more probable that this training example is $x_{veiled}$ -- an undetected veiled offense by the original compromised classifier $C$.
If this is indeed true, we may surface multiple $x_{veiled}$ instances in the top influential training examples, and enable training data corrections with a high efficiency.

\begin{figure}[t]
    \includegraphics[width=0.49\textwidth]{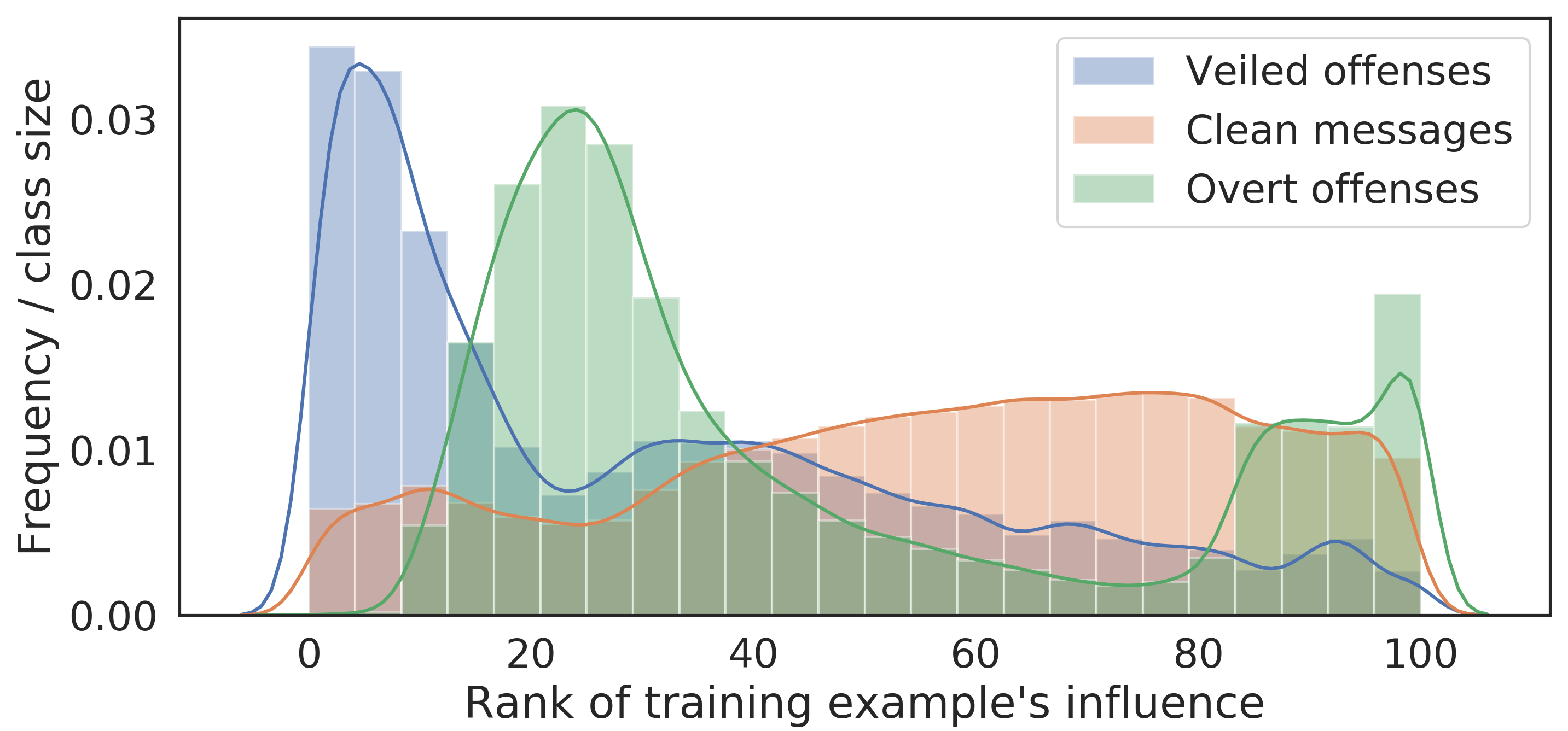}
    \caption{Distribution of each training example's influence rank to each probing veiled offense by the \emph{gradient product} method. An influence rank towards 0(\%) means a high influence.
    }
    \label{fig:trackin_dist_fig}
\end{figure}

\begin{table}[t]
\small
\begin{center}
\renewcommand{\arraystretch}{1.3}
\begin{tabular}{p{0.16\textwidth}rrrr}
    \toprule
    {Method} & @ 500 & 1000 & 1500 & 2000\\
    \midrule
    {Random} & 100 & 200 & 300 & 400\\
    {Training loss} & 267 & 464 & 634 & 775\\
    {Embedding product} & 162 & 303 & 445 & 570\\
    {Influence function} & 259 & 472 & 647 & 806\\
    {Gradient product} & \textbf{290} & \textbf{563} & \textbf{777} & \textbf{961}\\
    \bottomrule
\end{tabular} 
\end{center}
\caption{
Number of veiled offenses found in the highest averagely ranked (most influential) training examples over all probing veiled offenses by different methods.
}
\label{tab:n_correctly_identified}
\end{table}

\begin{table}[t]
\small
\begin{center}
\renewcommand{\arraystretch}{1.3}
\begin{tabular}{p{0.146\textwidth}p{0.097\textwidth}rrr}
    \toprule
    {Model} & Operation & VO & NO & OO\\
    \midrule
    {\underline{Original}} & & 1.2 & 99.6 & 97.2\\
    [4pt]
    {\underline{Training loss}} & {fix top 500} & 8.9 & 99.5 & 97.0\\
    & {fix top 2000} & 26.7 & 98.2 & 98.2\\
    & {flip top 500} & 13.0 & 97.9 & 98.2\\
    & {flip top 2000} & 41.5 & 85.2 & 99.4\\
    [4pt]
    {\underline{Embedding product}} & {fix top 500} & 3.4 & 99.7 & 95.6\\
    & {fix top 2000} & 16.9 & 98.7 & 98.6\\
    & {flip top 500} & 7.3 & 97.1 & 96.3\\
    & {flip top 2000} & 29.9 & 85.4 & 98.3\\
    [4pt]
    {\underline{Influence function}} & {fix top 500} & 9.5 & 99.5 & 97.3\\
    & {fix top 2000} & 28.0 & 98.7 & 98.0\\
    & {flip top 500} & 12.7 & 98.5 & 98.1\\
    & {flip top 2000} & 41.7 & 87.5 & 99.6\\
    [4pt]
    {\underline{Gradient product}} & {fix top 500} & 10.4 & 99.0 & 96.9\\
    & {fix top 2000} & 37.5 & 97.6 & 98.0\\
    & {flip top 500} & 14.6 & 97.6 & 98.4\\
    & {flip top 2000} & 51.1 & 87.6 & 99.5\\
    [4pt]
    {\underline{Gold}} & & 76.0 & 94.8 & 98.2\\
    \bottomrule
\end{tabular} 
\end{center}
\caption{
Class recall of the original model and the retrained models by different influence tracking methods on the veiled offensive (VO) test set, non-offensive (NO) test set, and overtly offensive (OO) test set.
}
\label{tab:performance}
\end{table}

In \autoref{fig:trackin_dist_fig}, we show a distribution of the training examples' influence rank derived by the gradient product method (\Sref{sec:background}).\footnote{Recall that each training example $x_{trn}$ has an influence score over each probing example $x_{prb}$. For each $x_{prb}$, we rank the influence of all $x_{trn}$, creating an \emph{influence rank} for each $x_{trn}$ in $\mathcal{D'}$ w.r.t.~$x_{prb}$.}
The influence ranks of the veiled offenses are highly skewed towards the left of the spectrum (more influential) compared to others, confirming our hypothesis. In \autoref{tab:n_correctly_identified}, we show the number of veiled offensive training examples surfaced among the most influential examples under different influence definitions. A random set of examples labeled as non-offensive would contain $\approx$20\% veiled offenses.\footnote{Since there are 2K veiled offenses and 8K non-offensive instances in the training set.} All probing methods attain 1.5--3x better results than random, with influence function and gradient product methods outperforming the training loss baseline that is often used in the active learning scenario.\footnote{We show in the Appendix that this result holds consistently even when probing with only 20 examples (instead of 100).}

Does the detection and re-annotation of veiled offenses improve the model? We retrain the original student classifier and for each influence metric we: (1) simulate the active learning scenario, fixing the labels for veiled offenses within the top $k$ examples (following gold SBIC annotations), and (2) flipping the labels for all of the top $k$ examples, including the non-offensive clean data. The former mimics a scenario where extra human annotation is available; the latter requires no extra human labor, and is motivated by the observation that crowdsourced annotations of veiled toxicity
are expensive and not always trustworthy.
But this setup might compromise the model performance on non-toxic data, since we will also flip innocuous influential training instances.

\autoref{tab:performance} shows the performance of the models improved by each method.
The \emph{gradient product} method achieves the best performance in recognizing the veiled offenses. Flipping the labels for all top influential training examples helps this performance on veiled offenses more than only correcting the labels, but also compromises the performance on non-offensive clean data.
All of the updated models still preserve the utility in recognizing the overt toxicity.
However, there remains a significant gap between our improved models and the benchmark model, which has the correct information of all veiled offenses, clean data and overtly offensive language in the training set.

\begin{table}[ht]
\small
\begin{center}
\renewcommand{\arraystretch}{1.3}
\begin{tabular}{p{0.153\textwidth}p{0.27\textwidth}}
    \toprule
    {Probing example} & {Literally the only thing that matters for choosing a wife - The woman's chastity.}\\
    \midrule
    {Training loss} & {What does Nike and KKK have in common? They both make the nigga’s run.}\\
    {Embedding product} & {Question: Can misandry be classified as sexual harassment and be subject to the same grounds for a lawsuit as misogyny?}\\
    {Influence function} & {I would pay money to hatefuck some of the female users on this sub.}\\
    {Gradient product} & {Of course I love you, I just prefer the straight part of you to the gay part.}\\
    \bottomrule
\end{tabular} 
\end{center}
\caption{
Top influential training examples to a probing veiled offense by different influence metrics.
}
\label{tab:actual_example}
\end{table}
\paragraph{Examples}
\autoref{tab:actual_example} shows a probing example and top influential training examples identified by each influence metric. Using the training loss metric, we get exactly the same influential examples for every probing instance since the metric does not depend on the probing examples.
It surfaces a racist comment, although Perspective API classified the whole sentence as not toxic.
For the influence function metric, although the surfaced misogynistic example is overtly offensive, it was not recognized as toxic by Perspective API as well.
The embedding product metric surfaces a post related to the topic of sexual harassment as the probing example, but the post is actually non-offensive, which underscores the need for further validation of surfaced comments using manual annotations or more sophisticated toxicity detection methods. Finally, the gradient product metric surfaces a homophobic microaggression. Overall, although these influence metrics can help finding candidates for veiled offensiveness, the discovered messages might not necessarily target the same social groups as the probing examples and also might not be toxic.
Future work should focus on incorporating knowledge about social groups to the classifiers, and on a deeper analysis of presuppositions encoded in the surfaced messages to analyze toxicity in conversational and social contexts.

\section{Conclusion}
We propose a framework to robustify toxicity classifiers against veiled toxicity. Through a few labeled probing examples, we can accurately surface orders of magnitude more disguised toxic messages missed by a compromised classifier, using interpretable ML techniques that track the influence of training examples on the probing examples.
Our framework, however, is not limited to toxicity detection. Future work can explore how to enhance a sub-optimal model using the teacher-student setup for tasks that change across domains or over time, or in scenarios where the original model and data are restricted for privacy reasons.

\paragraph{Acknowledgments.}
We thank the anonymous EMNLP reviewers and area chairs, Alan Black, Anjalie Field, Junxian He, Eden Tsvetkov, Jing Wen, Michael Miller Yoder, and members of TsvetShop at CMU for helpful discussions of this work.
This material is based upon work supported by NSF grants IIS1812327 and SES1926043, by the Public Interest Technology University Network Grant No.~NVF-PITU-Carnegie Mellon University-Subgrant-009246-2019-10-01, the Okawa Grant, and by Amazon MLRA award.
We also thank Amazon for providing GPU credits.

\bibliography{my_cites}
\bibliographystyle{acl_natbib}

\appendix
\section{Implementation Details}
The student classifier we used for experiments is a BERT-Base model \citep{Devlin2019BERTPO}, adapted from \citet{Wolf2019HuggingFacesTS}. The model has 110 million parameters. We used the default BERT optimizer with default hyperparameters: a learning rate of 5e-5, a total of 3 epochs, a max sequence length of 200, and a training batch size of 24. The training (finetuning) of the student classifier would take approximately 10 minutes on one NVIDIA GeForce RTX 2080 Ti GPU.

For influence functions, we followed \citet{Han2020ExplainingBB} which adapted code from \citet{Koh2017UnderstandingBP} to PyTorch and used the same stochastic estimation trick, LiSSA \citep{Agarwal2016SecondOrderSO}. Since our model is not convex, we used a ``damping'' term of 3e-3. This value was picked so that the recursive approximation to the inverse Hessian-vector product can be finished (converged) in a reasonable time.\footnote{15 minutes for each training and probing example pair on one NVIDIA GeForce RTX 2080 Ti GPU.} More specifically, we chose the recursion depth to be 3000 (with a total of 12k training examples), the number of recursions to be 1, and a scaling factor to be 1e8. In each step estimating the Hessian-vector product, we took a batch of 8 training examples for stability. We empirically checked that the inverse Hessian-vector product converges after the recursive estimation for all probing examples on which we performed the analysis.

For the gradient product method, we adopted \emph{TrackInCP (Equally Weighted)} from \citet{Pruthi2020EstimatingTD}. We used 3 checkpoints of the BERT model (i.e., $k=3$; one checkpoint after each epoch) since we finetuned the classifier with 3 epochs.

\section{Sample Veiled Offenses}
\autoref{tab:random_probing_examples} shows 20 random veiled offensive examples as extracted in \Sref{sec:model_and_data_instantiation}. They got low toxicity scores from Perspective API, but were annotated as offensive to at least one social group according to the SBIC dataset. Most of them are \emph{covert} but no less offensive to human readers.

\begin{table*}[ht]
\small
\begin{center}
\renewcommand{\arraystretch}{1.3}
\begin{tabular}{p{0.73\textwidth}p{0.21\textwidth}}
    \toprule
    {Original text} & {Targeted group annotation}\\
    \midrule
{Now we'll see if this dude is everything that he's cracked up to be. Time to get ready to bring the Travel Ban up to the Supreme Court.} & {Muslim folks}\\
{How does a Rabbi get paid? He keeps the tips} & {Jewish folks}\\
{if a man speaks and his wife is n't around to correct him .  is he still wrong ?} & {Women}\\
{Both are important, but being Neurotypical is more important than good looks, women won't approach you unless you're a 10/10 god cope all you want, if you're decent/good-looking (6-8/10) women won't approach you} & {Women}\\
{what is the difference between a bench and an african ? a bench can support a family .} & {Black folks}\\
{I recently have trying to eat more vegetables to lower my blood pressure. The hospital security has really been making this impossible lately though. Guess ill have to just try the morgue.} & {Physically disabled folks}\\
{How do you keep your vegetables happy? Gather them on an air-conditioned room with plenty of sunlight} & {Physically disabled folks}\\
{What's the difference between boy scouts and jews? Boy scouts come back from camp.} & {Jewish folks}\\
{What’s blue and white and red all over? A cop at a BLM rally.} & {Black folks}\\
{I walk into the bank the other day, I'm already having a bad day, and the teller gives me the look, I know this look. Teller:: What are you? me:: Why? teller:: Well, you look Middle-Eastern, but then I saw your name, and I thought that, that couldn't be right! me:: What makes you think I'm Middle-Eastern - my beard, my scarf, my skin color? Teller:: um... This kind of thing happens to be pretty much every single day, no matter where I (half Mexican) am.} & {Middle eastern}\\
{I don't get why Black Friday is such a big deal. They already have a whole history month} & {Black folks}\\
{There’s a lot of wildfires around us and I noticed some ashes falling down at my Jewish friend’s house I called to make sure his family reunion was going well} & {Jewish folks}\\
{If Donald Glover played spiderman, he wouldn't be called spiderman anymore. He would be called Spidermonkey.} & {Black folks}\\
{But he can never be a *real* girl.} & {Trans women}\\
{For your viewing pleasure, a full album of defending feminism in this subreddit, and then getting banned from the feminism subreddit} & {Women}\\
{I want to have 13 children with a proud Aryan woman.} & {Any non white person}\\
{What’s a preist’s favourite song Praising the lord in A minor} & {Christian folks}\\
{Life is like a box of chocolates The ones with nuts are generally better} & {Women}\\
{Why do deaf children not get abused Because there’s no need to give them beats.} & {Physically disabled folks}\\
{I'm not judgemental So whenever I see someone driving slow in the fast lane, I never assume what gender she is.} & {Women}\\
    \bottomrule
\end{tabular} 
\end{center}
\caption{
Sample veiled offenses extracted from SBIC.
}
\label{tab:random_probing_examples}
\end{table*}

\section{Robustness Check}
In the main paper, we used 100 labeled veiled offensive examples as our probing set to discover more veiled offenses in the training set. As a robustness check, we show in \autoref{tab:_n_correctly_identified_20} that the result is consistent when using only 20 probing examples.

\begin{table*}[t]
\small
\begin{center}
\renewcommand{\arraystretch}{1.3}
\begin{tabular}{p{0.16\textwidth}rrrr}
    \toprule
    {Method} & @ 500 & 1000 & 1500 & 2000\\
    \midrule
    {Random} & 100 ($\pm$ 0) & 200 ($\pm$ 0) & 300 ($\pm$ 0) & 400 ($\pm$ 0)\\
    {Training loss} & 267 ($\pm$ 0) & 464 ($\pm$ 0) & 634 ($\pm$ 0) & 775 ($\pm$ 0)\\
    {Embedding product} & 161.2 ($\pm$ 19.5) & 305.8 ($\pm$ 19.4) & 442.8 ($\pm$ 35.9) & 567.0 ($\pm$ 44.5)\\
    {Influence function} & 258.8 ($\pm$ 1.6) & 469.6 ($\pm$ 2.9) & 645.6 ($\pm$ 2.1) & 807.4 ($\pm$ 2.7)\\
    {Gradient product} & \textbf{289.2} ($\pm$ 4.3) & \textbf{563.6} ($\pm$ 3.5) & \textbf{775.2} ($\pm$ 1.6) & \textbf{962.0} ($\pm$ 4.4)\\
    \bottomrule
\end{tabular} 
\end{center}
\caption{
Number of veiled offenses found in the top influential training examples to the probing veiled offenses by different methods. We use 20 probing examples each time and repeat the experiment 5 times. The result shows both the mean number and the 95\% confidence interval under $t$-distributions.
}
\label{tab:_n_correctly_identified_20}
\end{table*}

\end{document}